\documentclass[conference]{IEEEtran}
\IEEEoverridecommandlockouts
\usepackage{cite}
\usepackage{amsmath,amssymb,amsfonts}
\usepackage{graphicx}
\usepackage{textcomp}
\usepackage{xcolor}
\usepackage{bm}
\usepackage{amsmath}
\usepackage{amssymb}
\usepackage{amsthm}
\usepackage{mathrsfs}
\usepackage{enumerate}
\usepackage{multirow}
\usepackage{color}
\usepackage{threeparttable}
\usepackage{subfigure}
\usepackage{tabularx}
\usepackage{caption}
\usepackage[square, comma, sort&compress, numbers]{natbib}
\usepackage[pagebackref=false,breaklinks=true,letterpaper=true,colorlinks,bookmarks=false]{hyperref}
\usepackage{times}
\usepackage{breakurl}
\usepackage{array}
\usepackage{verbatim}
\usepackage{algorithm}
\usepackage{bbding}
\usepackage{algpseudocode}
\usepackage{lettrine}
\usepackage{svg}
\usepackage{color}
\usepackage{multirow}
\usepackage{graphicx}
\usepackage{caption}
\usepackage{float}
\usepackage{pifont}
\usepackage{caption}
\usepackage{float} 

\usepackage{booktabs}

\def\BibTeX{{\rm B\kern-.05em{\sc i\kern-.025em b}\kern-.08em
    T\kern-.1667em\lower.7ex\hbox{E}\kern-.125emX}}
\begin{document}

\title{LR-FPN: Enhancing Remote Sensing Object Detection with Location Refined Feature Pyramid Network\\
{\footnotesize 
}
}

\author{
  Hanqian Li\textsuperscript{1}, ~~Ruinan Zhang\textsuperscript{1},  ~Ye Pan\textsuperscript{2}, ~Junchi Ren\textsuperscript{3},  ~Fei Shen\textsuperscript{4 5 *}\thanks{* Corresponding author} \\
  \textsuperscript{1}Shandong University, Jinan, China \\
  \textsuperscript{2}South China Normal University, Foshan, China\\
  \textsuperscript{3}China Telecom Corporation Limited, Nanjing, China\\
  \textsuperscript{4}Nanjing University of Science and Technology, Nanjing, China \\
  \textsuperscript{5}Tencent AI Lab, Shenzhen, China
}

\maketitle
\begin{abstract}
Remote sensing target detection aims to identify and locate critical targets within remote sensing images, finding extensive applications in agriculture and urban planning. 
Feature pyramid networks (FPNs) are commonly used to extract multi-scale features. 
However, existing FPNs often overlook extracting low-level positional information and fine-grained context interaction.
To address this, we propose a novel location refined feature pyramid network (LR-FPN) to enhance the extraction of shallow positional information and facilitate fine-grained context interaction. 
The LR-FPN consists of two primary modules: the shallow position information extraction module (SPIEM) and the contextual interaction module (CIM). 
Specifically, SPIEM first maximizes the retention of solid location information of the target by simultaneously extracting positional and saliency information from the low-level feature map. 
Subsequently, CIM injects this robust location information into different layers of the original FPN through spatial and channel interaction, explicitly enhancing the object area.
Moreover, in spatial interaction, we introduce a simple local and non-local interaction strategy to learn and retain the saliency information of the object. 
Lastly, the LR-FPN can be readily integrated into common object detection frameworks to improve performance significantly.
Extensive experiments on two large-scale remote sensing datasets (i.e., DOTAV1.0 and HRSC2016) demonstrate that the proposed LR-FPN is superior to state-of-the-art object detection approaches.
Our code and models will be publicly available.

\end{abstract}


\section{Introduction}
Remote sensing object detection \cite{CMSCGC, R3det, FPN, hu2023bag} involves detecting and pinpointing objects within remotely captured images. It is a critical tool across multiple domains, notably in agriculture and urban planning. Efficiently and accurately identifying objects in remote sensing images is pivotal for informed decision-making, fostering sustainable development, and optimizing resource utilization. 

The feature pyramid network (FPN) \cite{FPN} stands as a prevalent model architecture widely employed in object detection tasks to construct a hierarchical feature representation akin to a pyramid structure. Its efficacy lies in propagating semantically robust features from higher to lower levels through a top-down approach and lateral connections. Despite the excellent performance demonstrated by FPN \cite{FPN} when it is proposed, there are still some flaws. Notably, certain design flaws within the feature pyramid network (FPN) are identified.
Flaws, as described, are in the following: \textbf{there is neglect in extracting and utilizing shallow localization information.}. 
The shallow layers of a backbone network usually contain valuable and exploitable localization information, but these pieces of information are often not effectively utilized or overlooked during the process of constructing a pyramid.
\textbf{There is a lack of interaction with contextual information.} 
Employing a $1\times1$ convolution directly before feature fusion fails to fully exploit interaction within the context,  potentially leading to a lack of interaction between the information and an inability to fully utilize the injected information. 
Addressing these issues is pivotal to prevent potential decreases in model accuracy.
\begin{figure}[t]
    \centering
    \includegraphics[width=1\linewidth]{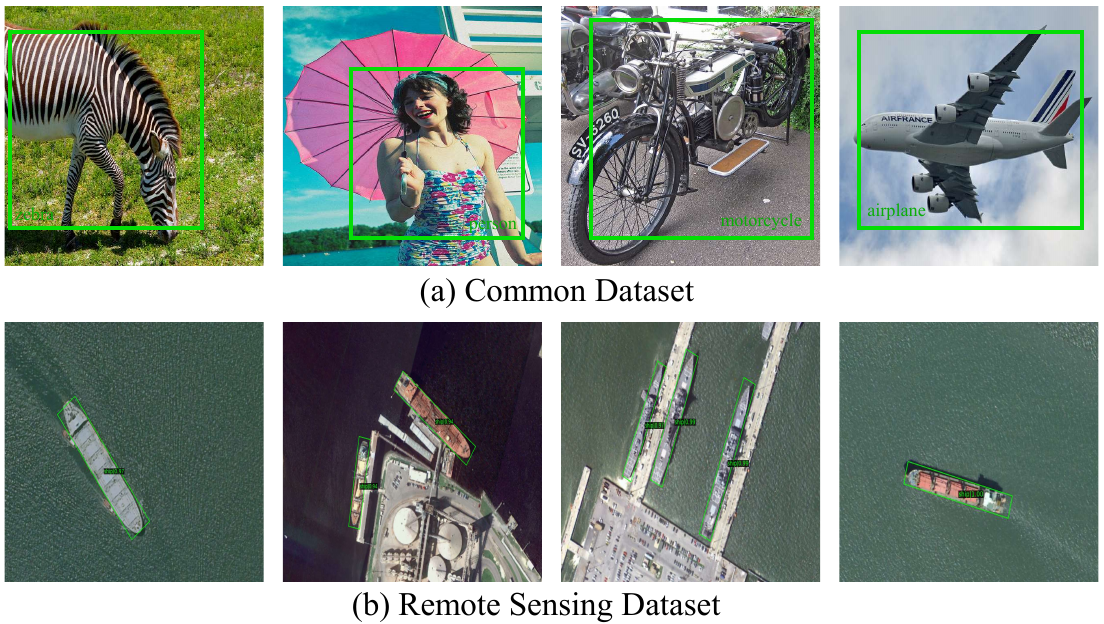}
\caption{Common dataset and remote sensing dataset. In typical datasets, target objects are larger, while in remote sensing datasets, they are comparatively smaller.}
\label{fig:dataset}
\end{figure}

\begin{figure*}[t]
    \centering
    \includegraphics[width=0.9\linewidth]{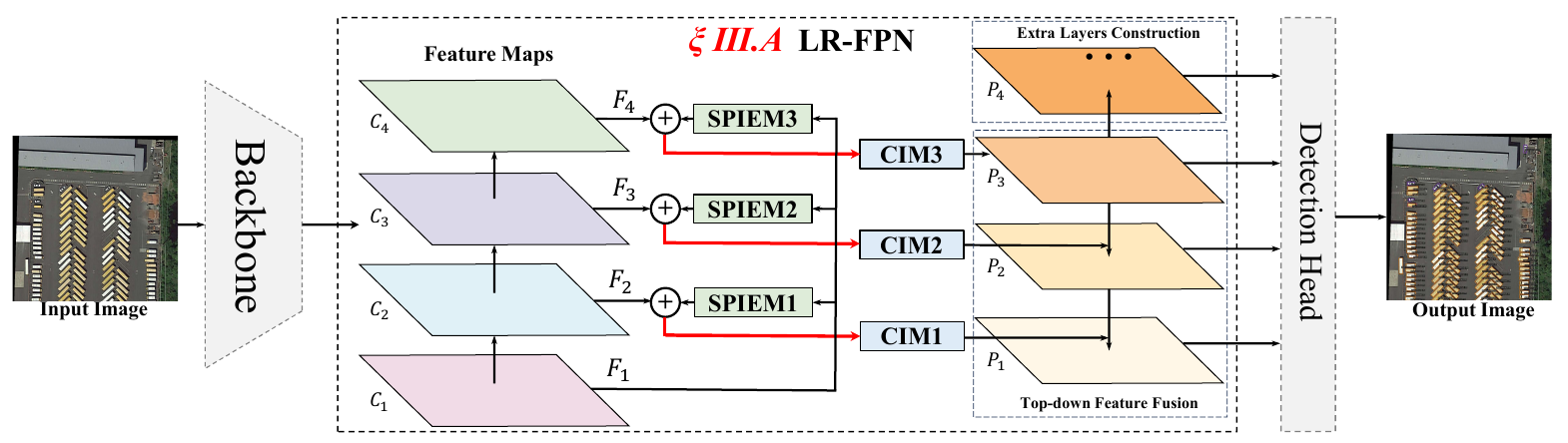}
\caption{ The architecture of our detector. The outputs of the individual shallow position information extraction module (SPIEM) are designed to adaptively align with the scales and channels of the feature maps in the backbone network. In the construction of the extra layers, we leverage the $3\times3$ convolution network for execution.}
\label{fig:overview}
\end{figure*} 

Recent advancements in models like AugFPN~\cite{AUGFPN} and PANet~\cite{pafpn} highlight the importance of enhanced feature fusion for accuracy. AugFPN ensures similar semantic information in feature maps post lateral connection and minimizes information loss in high-level features. PANet uses precise localization cues to shorten the information pathway and directs key information to proposal subnetworks via adaptive feature pooling.
NAS-FPN, beyond manual fusion strategies, uses neural architecture search methodologies for architecture optimization, improving various backbone models.
However, these FPNs are not specifically designed for remote sensing scenes, which typically have a single viewpoint and high density, as shown in Fig. \ref{fig:dataset}. For remote sensing applications, the network's ability to extract object location information and facilitate contextual interactions is crucial. Existing FPNs often overlook low-level positional information and fine-grained context interaction extraction, limiting their performance in remote sensing scenes.

 \begin{figure*}[t]
    \centering
    \includegraphics[width=0.9\linewidth]{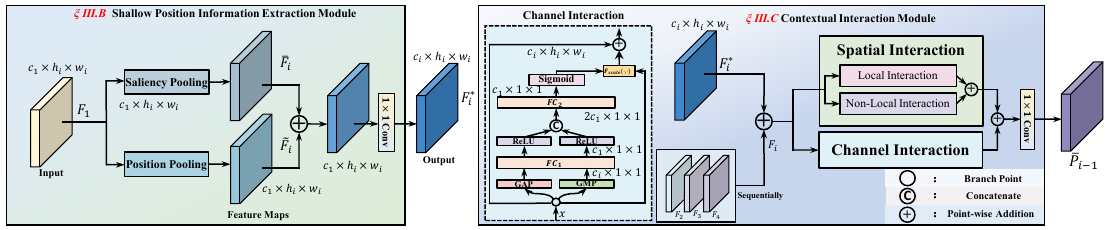}
\caption{The structure of shallow position information extraction module (SPIEM) and contextual interaction module (CIM).  CIM, GAP and GMP represent the global average and max pooling, respectively.}
\label{fig:mouduels}
\end{figure*} 

This paper introduces a novel Location Refined Feature Pyramid Network (LR-FPN) to enhance shallow positional information extraction and facilitate fine-grained context interaction. LR-FPN refines location information within the feature pyramid, compensating for location and saliency information across layers, and boosting contextual interaction. This maximizes the use of robustly extracted location information, improving task performance.
This refinement is achieved through two key modules: the Shallow Position Information Extraction Module (SPIEM) and the Context Interaction Module (CIM). SPIEM effectively adjusts position information, capturing positional and saliency details from low-level feature maps to maintain accurate target location information, facilitating compensation between layers.
Complementarily, CIM infuses reliable location information into different FPN layers through spatial and channel interaction, enhancing the object area. This process bolsters contextual information interaction, seamlessly integrating location information into the FPN and improving layer interaction, thereby maximizing information utilization.

The main contributions of this paper can be summarized
as follows: 

\begin{itemize}
\item  This paper presents a plug-and-play location refined feature pyramid network (LR-FPN) to enhance the extraction of shallow positional information and facilitate fine-grained context interaction.

\item  We present the shallow position information extraction module (SPIEM) and the context interaction module (CIM) to extract positional and saliency information from the low-level feature maps, thereby maximizing the enhancement and retention of location information. Furthermore, they facilitate interaction with other layers in both spatial and channel dimensions.

\item We conduct extensive experiments and achieve promising performance gains on two large-scale object datasets. 
Besides, the ablation studies also verify the effectiveness of the core mechanisms in the LR-FPN for object detection in remote sensing.

\end{itemize}

\section{Related Work}\label{sec:rw}

\subsection{Remote Sensing Object Detection}  
{\color{black} In recent years, object detection~\cite{weng2023cross,weng2024enhancing,qiao2022novel} has been gaining increasing popularity in the computer vision community. AAF-Faster RCNN~\cite{cascade} introduces the additive activation function, aiming to enhance the detection accuracy of small-scale targets. CF2PN~\cite{cf2pn} incorporates multi-level feature fusion techniques to address the challenge of inefficient detection of multi-scale objects in remote sensing object detection tasks. RICA~\cite{li2017rotation} introduces an RPN with multi-angle anchors into Faster R-CNN to tackle remote sensing targets with arbitrary orientations.  While these methods have to some extent enhanced the accuracy of remote sensing object detection, they don't take into account the integration of location information with spatial and channel information to improve the performance of remote sensing object detection.}

\subsection{Context Exploitation}  
The extraction of contextual information is widely utilized across various domains \cite{liu2019large,lai2023shared, li2022enhancing,lai2023faithful} within artificial intelligence. 
For example,
Deeplab-v3 \cite{deeplabv3} utilizes atrous convolution which magnifies the receptive field to acquire muti-scale context while decreasing the loss of information. HPGN \cite{shen2021exploring} proposes a novel pyramid graph network targeting features, which is closely connected behind the backbone network to explore multi-scale spatial structural features. PBSL \cite{shen2023pbsl} introduces a bidirectional perception module to explore the contextual relationships of prominent features between positive and negative samples.
PSP-Net \cite{pspnet} makes use of the pooling to extract the hierarchical global context. However, the bottom feature layer often contains valuable positional information that is sometimes overlooked.

\subsection{{Multi-scale Feature Fusion}} 
Multi-scale features possess the capability to capture object representations at various scales in images, showcasing excellent performance across multiple tasks \cite{lai2023multimodal, shen2023git, xu2023cross, 2022ResCapsNet, PSCPC, liu2023adaptive, 2024AMGC}.
NAS-FPN \cite{nasfpn} employs reinforcement learning to train a controller that identifies the optimal model architectures within a predefined search space. EMRN \cite{emrn} proposes a multi-resolution features dimension uniform module to fix dimensional features from images of varying resolutions. BiFPN \cite{bifpn} removes those nodes that only have one input edge, adds an extra edge from the original input to the output node if they are at the same level, and treats each bidirectional path as one feature network layer, which makes better accuracy and efficiency trade-offs. However, these methods fail to consider the extraction of low-level location information and the nuanced interaction within the context.

\section{Proposed Method}\label{sec:method}
\subsection{Overview}
Based on the original structure of FPN \cite{FPN}, we denote the feature maps used to build the feature pyramid as \{F$_1$, F$_2$, F$_3$, F$_4$\}, and denote the outputs of the feature pyramid as \{P$_1$, P$_2$, P$_3$, P$_4$, P$_5$\}. The structure is shown in Fig. \ref{fig:overview}.
In the construction of the feature pyramid, we add extra layers by using $3\times3$ convolutions to expand the information from \{P3\} to \{P4, P5\}, which will ultimately provide 5 scales of outputs for subsequent detection heads. The overall computation formula of the network we designed is as follows:
\begin{equation}
    P_{3} = f_{4}(F_{4}+F_{4}^{*}),
\end{equation}
where $P$ denotes the outputs in our FPN. $f_{i}(\cdot)$ denotes the lateral connection block we redesign, which will be introduced later. $F$ denotes the inputs gained from the backbone. $F_{1}^{*}$ denotes the outputs of the shallow position information extraction operation, which is introduced later. 
\begin{equation}
    \overline P_{i-1} = f_{i}(F_{i}+F_{i}^{*} ),i=2,3,
\end{equation}
where $i$ represents the layer of the pyramid. 
\begin{equation}
    P_{i-1}=\overline P_{i-1}+R(P_{i})
\end{equation}
The operation denoted by $R(\cdot)$ is used to resize the features so that they have the same spatial sizes. 
\begin{equation}
     P_{k+1} = Conv_{3\times3}(P_{k}),k=3,4,
\end{equation}
where $Conv_{3\times3}$ denotes the convolutional neural network with the kernel size of 3. Two components of our FPN will be described in the following subsections.
\begin{table*}[htbp]
  \caption{SOTA Comparison on DOTAV1.0 \cite{DOTA_dataset}.  The abbreviations of categories are defined as: Plane (PL), Baseball diamond (BD), Bridge (BR), Ground field track (GTF), Small vehicle (SV), Large vehicle (LV), Ship (SH), Tennis court (TC), Basketball court (BC), Storage tank (ST), Soccer-ball field (SBF), Roundabout (RA), Harbor (HA), Swimming pool (SP), and Helicopter (HC). We respectively bold and underline the optimal and suboptimal results for each category metric.}
    \label{tab:sota}
  \renewcommand\arraystretch{1.1} 
  \centering\resizebox{0.98\textwidth}{!}{
  \begin{tabular}{*{17}{l|}l}
    \toprule
    \centering
    \textnormal{Method} & \textnormal{Backbone}& \textnormal{mAP} & \textnormal{PL} & \textnormal{BD} & \textnormal{BR} & \textnormal{GTF} & \textnormal{SV} & \textnormal{LV} & \textnormal{SH} & \textnormal{TC} & \textnormal{BC} & \textnormal{ST} & \textnormal{SBF} & \textnormal{RA} & \textnormal{HA} & \textnormal{SP} & \textnormal{HC} \\
    \midrule
    \textnormal{IENet \cite{ienet}} & \color{gray}{ResNet101}& 57.14 & 80.20 & 64.54 &  39.82 &  32.07 &  49.71 &  65.01 &  52.58 &  81.45 &  44.66 &  78.51 &  46.54 &  56.73 &  64.40  & 64.24  & 36.75  \\
     \textnormal{R-DFPN \cite{rd-fpn}}	 & \color{gray}{ResNet101}&57.94&80.92&65.82&33.77&58.94&55.77&50.94&54.78&90.33&66.34&68.66&48.73&51.76&55.10&51.32&35.88\\
    \textnormal{PIoU \cite{piou}} & DLA-34 & 60.53  &   80.91 &  69.70 &  24.11 &  60.22  & 38.34 &  64.43  & 64.84 &  \textbf{90.98}  & 77.22 &  70.45 &  46.59 &37.12&  57.11 &  61.97 &  \textbf{64.02} \\
    \textnormal{Faster-RCNN \cite{Faster-rcnn}} & ResNet50&  60.42  &  80.31 &  \textbf{77.54} &  32.84 &  68.11  & 53.67 &  52.44  & 50.05 &  90.45  & 75.05 &  59.55 &56.09 &  49.81 &  61.67 &  56.42 &  41.87 \\
    \textnormal{Light-Head R-CNN \cite{lighter_head_rcnn}} & \color{gray}{ResNet101}& 66.95& 88.02 & 76.99 & 36.70 & \underline{72.54} & 70.15 & 61.79 & 75.77 & 90.14 & 73.81 & \underline{85.04} & 56.57 & 62.63 &  53.30 & 59.54 & 41.91 \\
    \textnormal{RADet \cite{Radet}} &\color{gray}{ResNet101}& 67.66& 79.66 & \underline{77.36} & \underline{47.64} & 67.61 & 65.06 & \underline{74.35} & 68.82 & 90.05 & 74.72 & 75.67 & 45.60 & 61.84 & 64.88 & \underline{68.00} & 53.67  \\
    \textnormal{H2RBox \cite{H2RBox}} & ResNet50 &67.82 &88.51 &73.52 &40.83 &56.91 &\underline{77.53}& 65.45 &77.93 &\underline{90.91} &\textbf{83.25} &\textbf{85.33} &55.31 &62.92 &52.45 &63.67 &43.38  \\    
    \textnormal{ICN \cite{ICAN}} & \color{gray}{ResNet101} & 68.16 &  81.36 & 74.30 & \textbf{47.70} & 70.32 & 64.89 & 67.82 & 69.98 & 90.76 & 79.06 & 78.20 & 53.64 & 62.90 & \textbf{67.02} & 64.17 & 50.23 \\
    \textnormal{GSDet \cite{gsdet}} & \color{gray}{ResNet101}& 68.34&  81.11 &76.85& 40.81&\textbf{75.93} &  64.57 &   58.43 &  74.24 &  89.93 &  79.43 &  78.83 &   \textbf{64.58} &  \underline{63.42} &  \underline{66.03}    & 58.01 & 52.24 \\
    \textnormal{PVANet-FFN \cite{PVANet-FFN}} &  PVANet&  \underline{69.08} &  \textbf{90.21} &  73.25 &  45.47 &  62.36  & 67.12 &  70.13  & 73.64 &  90.38  & 76.58 &  80.45 &  49.59 &  59.61 &  65.87 &  \textbf{68.62} &  \underline{62.97}\\
    \hline
    R$^3$Det \cite{R3det} (Baseline) & \color{gray}{ResNet101}&67.54&89.41 &73.10 &40.54& 57.48 &76.48 &72.32 &\underline{78.99} &90.88 &76.71 &83.49 &52.89& 60.77& 61.29& 59.45 &39.40\\
    \textnormal{LR-FPN} (Ours) & ResNet50& \textbf{69.68}& \underline{89.46} &74.53& 42.46&64.88 &  \textbf{77.90} &   \textbf{75.48} &  \textbf{83.96} &  \textbf{90.98} &  \underline{81.33} &  83.69 &  \underline{56.58} &  \textbf{63.46} & 65.36& 60.89 & 36.29\\
    \bottomrule    
\end{tabular} }
\end{table*}

\begin{table}[tb!]
		\centering
  \caption{SOTA Comparison on HRSC2016 \cite{HRSC2016}.}
		{
			\begin{tabular}{l|c|c}
				\toprule
				Method & Backbone & mAP \\
				\midrule
                R$^2$CNN \cite{r2cnn} & \color{gray}{ResNet101} & 73.1  \\
				RC1 \& RC2 \cite{r1_r2} & VGG16 & 75.7   \\
				RRPN \cite{rrpn} & \color{gray}{ResNet101}  & 79.1  \\
				R$^2$PN \cite{r2pn}  &  VGG16 & 79.6  \\
				RetinaNet-H & \color{gray}{ResNet101} & 82.9  \\
				RRD \cite{rdd} & VGG16  & 84.3 \\
				RoI-Transformer \cite{roi_transformer} & \color{gray}{ResNet101} & 86.2  \\
				Gliding Vertex \cite{Gliding_Vertex} &\color{gray}{ResNet101} &  88.2  \\
                AFPN \cite{AFPN} & \color{gray}{ResNet101}& \underline{89.4} \\
				\hline
                     R$^3$Det \cite{R3det} (Baseline) & \color{gray}{ResNet101} & 88.2\\
                LR-FPN (Ours) & ResNet50  & \textbf{90.4} \\
			\bottomrule
		\end{tabular}}
		\label{table:HRSC2016}
	\end{table}
\subsection{Shallow Position Information Extraction Module}  
FPN \cite{FPN} leverages the feature hierarchy within a neural network to generate feature maps with varying resolutions. However, the features that target the improvement of semantic information lose the position information from the shallow layer, decreasing the positioning function in high layers. What's more, in remote sensing detection tasks, similar objects are always in dense arrangement, which needs more powerful capability in positioning.

This inspires us to propose location extraction fusion, namely the shallow position information extraction module (SPIEM). In this module, we expect to supply the location information before the fusion. Here is the computation formula for this module:
\begin{equation}
    \overline{F}_{i} = \overline{W}_i \odot  AAP_i(F_{1}),
\end{equation}
\begin{equation}
AAP(F^k)=\frac{1}{h\times w} \sum_{i=1}^{h} \sum_{j=1}^{w}F_{i,j}^k,
\end{equation}
where $\overline{W}_i$ denotes the adaptive weight. $\odot$ denotes the hadamard product. $AAP(\cdot)$ denotes the adaptive average pooling layer to adapt different scales in \{C$_2$, C$_3$, C$_4$\}.
\begin{equation}
   \widetilde{F}_{i} = \widetilde{W}_i \odot AMP_{i}(F_{1}),
\end{equation}
\begin{equation}
   AMP(F^k)= \underset{i=1,2,...,h}{max} \quad \underset{j=1,2,...,w}{max} (F_{i,j}^k ),
\end{equation}
where $\widetilde{W}_i$ denotes the adaptive weight. $AMP(\cdot)$  denotes the adaptive max pooling layer.
\begin{equation}
F_{i}^{*} = Conv_{1\times1}({\overline{F}{_i}+\widetilde{F}{_i}}),
\end{equation}
where $Conv_{1\times1}$ denotes the convolutional neural network with the kernel size of 1.

 \begin{table*}[t!]
\centering
\caption{Ablation study on key modules of LR-FPN.
The shallow position information extraction module (SPIEM) has two critical components: the saliency pooling
 ($\bf{SP}$) and the position pooling
 ($\bf{PP}$). The contextual interaction module (CIM) has two essential components: the spatial interaction
 ($\bf{SI}$) and the channel interaction
 ($\bf{CI}$).
}

\begin{tabular}{c||cc|cc|cc|ccc}
\hline
  \hline
   \multirow{3}*{No.} & \multicolumn{2}{c|}{\multirow{3}*{Methods}} & \multicolumn{4}{c|}{Components} & \multicolumn{3}{c}{Metric}  \\
  \cline{4-10}
  &  & &\multicolumn{2}{c|}{SPIEM}  &\multicolumn{2}{c|}{CIM}
  & \multirow{2}*{AP$_{50}$}  & \multirow{2}*{AP$_{75}$} & \multicolumn{1}{c}{\multirow{2}*{mAP}}  \\
  \cline{4-7}
  & & & $\bf{SP}$ & $\bf{PP}$ & $\bf{SI}$  & $\bf{CI}$  &\multicolumn{3}{c}{}\\
  \hline
  0 & \multicolumn{2}{c|}{Baseline}  &\color{gray}\ding{55}    &\color{gray}\ding{55}   &\color{gray}\ding{55}   &\color{gray}\ding{55} & 88.2 & 65.4 & 55.9 \\
  1 & \multicolumn{2}{c|}{+ SPIEM} &\ding{51}    &\ding{51}       &\color{gray}\ding{55}   &\color{gray}\ding{55}         & 89.1 & 66.3 & 56.1 \\
  2 & \multicolumn{2}{c|}{+ CIM}  &\color{gray}\ding{55}    &\color{gray}\ding{55}  &\ding{51}     &\ding{51} & 89.7 & 66.1 & 56.2  \\
  \hline
  3 & \multicolumn{2}{c|}{+ CIM + SPIEM (only $\bf{SP}$))}    &\ding{51}  &\color{gray}\ding{55}  & \ding{51}    & \ding{51}    & 89.7 & 67.6 & 56.3 \\
4 & \multicolumn{2}{c|}{+ CIM + SPIEM (only $\bf{PP}$))}    &\color{gray}\ding{55}  &\ding{51}  & \ding{51}    & \ding{51}    & 89.9 & 67.0 & 56.3 \\
  5 & \multicolumn{2}{c|}{+ SPIEM + CIM (only $\bf{SI}$))}        &\ding{51}   &\ding{51}     & \ding{51} &\color{gray}\ding{55} & 89.7 &66.7   & 57.2 \\
  6 & \multicolumn{2}{c|}{+ SPIEM + CIM (only $\bf{CI}$))}        &\ding{51}  &\ding{51}     &\color{gray}\ding{55} &  \ding{51}   & 89.8 & 67.0 &57.2 \\
    \hline
  7 &  \multicolumn{2}{c|}{LR-FPN}      &\ding{51}      &\ding{51}     &\ding{51}     &\ding{51}    &{\bfseries 90.4} &{\bfseries 68.9} &{\bfseries 59.5}\\
  \hline
    \hline
\end{tabular}
\label{tab:hrsc ablation}
\end{table*}
Initially, we construct a feature pyramid by utilizing the multi-scale features as inputs obtained from the backbone layers (\{C$1$, C$_2$, C$_3$, C$_4$\}). Then our FPN outputs the aggregated features \{P$_1$, P$_2$, P$_3$, P$_4$, P$_5$\}. 
We hope to reduce the computational cost of operation so we don't use the feature \{F$_1$\} to build a lateral connection because of the large scale. However, We extract localization information from the feature \{F$_1$\} through the module, and then append this information to \{F$_2$, F$_3$, F$_4$\} before the lateral connection operation. Considering some salient information in \{F$_1$\}, we utilize saliency pooling to increase saliency. SPIEM optimizes the preservation of precise target location details while simultaneously extracting positional and saliency information.

\subsection{Contextual Interaction Module} 
Original FPN  does not efficiently interact with spatial and channel information. To address this, we introduce an enhanced lateral connection module, the contextual interaction module (CIM), designed to tackle this issue more effectively. We separate the channel and spatial information with the detailed framework illustrated in Fig. \ref{fig:mouduels}. We substitute the $1 \times 1$ convolution network in the original FPN with a combination of depthwise convolutions, dilated depthwise convolutions, and a network of channel interaction modules in a parallel-addition form.
Our central concept is to utilize depthwise convolution to interact with local spatial information within each channel. Concurrently, we employ dilated depthwise convolutions to tackle the challenge of non-local spatial interaction. This approach expands the receptive field and enhances the non-local interaction of spatial information, thereby addressing the limitations of the original FPN structure.

In the channel interaction branch, we prioritize the interplay of information between channels. To harness key information and further boost the overall performance of lateral connections, we introduce a weighted processing method between channels. This method allows us to model the significance of different channels and adjust the contribution of information accordingly, thereby strengthening the lateral connections. To encapsulate the global information and key features of each channel, we employ global average and max pooling techniques respectively. Following this, we use a fully connected network, FC$_1$, to jointly learn the shared weights of these two types of information. The results are then concatenated and passed through another fully connected network,FC$_2$, and corresponding activation functions to derive the weights of each channel. These weights are subsequently applied to the original channels.
The final output is obtained through residual connections, ensuring the preservation of original information while adding the weighted enhancements. The computation formula of this specially designed module is as follows:

\vspace{-0.1cm}
\begin{equation}
    A_{i}(x) = Wx+x,i=2,3,4,
\end{equation}
\vspace{-0.1cm}
where $A_{i}$ denotes the channel interaction operation we design and $W$ denotes the weight generated by the channel interaction block.
\begin{equation}
\widetilde{f_{i}}(x) = Conv_{DW}(x) + Conv^{*}_{DW}(x) + A_{i}(x),
\end{equation}
where $Conv_{DW}$ denotes the depthwise convolution. $Conv^{*}_{DW}$ denotes the dilated depthwise convolution.
\begin{equation}
    f_{i}(x) = Conv_{1\times1}(\widetilde{f_{i}}(x)), i=2,3,4,
\end{equation}
where $ Conv_{1\times1}$ denotes the convolution neural network with the kernel size of 1.

In comparison to the original $1\times1$ convolution network in FPN \cite{FPN}, Our CIM leverages spatial and channel interaction to effectively incorporate robust location information into various layers of the original FPN, explicitly enhancing the representation of object areas.

\section{Experiment and Analysis}\label{sec:exp} 
\subsection{Datasets}
\textbf{\emph{DOTAV1.0}} \cite{DOTA_dataset} dataset comprises 2,806 high-resolution aerial images captured by various sensors and platforms.  Due to the inclusion of angle information and relatively small object sizes in DOTAV1.0 
 \cite{DOTA_dataset} datasets, each image is divided into $1024 \times 1024$ subimages, with an overlap of 200 pixels to make the detection task smooth.

\textbf{\emph{HRSC2016}} \cite{HRSC2016} dataset comprises images from two scenarios, encompassing ships at sea and ships close to the shore. The images are collected from six renowned harbors. The image dimensions range from $300\times300$ to $1,500\times900$. The dataset is divided into two sets: a training set with 436 images and a testing set with 444 images.

\subsection{Implementation Details} 
All experiments are conducted using MMRotate as the underlying framework and developed by PyTorch. To show the feasibility and efficiency of LR-FPN, we choose ResNet50 as the backbone instead of ResNet101. Then, we resize the image into (1024,1024) in DOTAV1.0 \cite{DOTA_dataset} and HRSC2016 \cite{HRSC2016}. We train the model using 1 GPU (RTX 3090) and set the batch size to 4. The learning rate initially set to 0.0025 is employed with a linear warm-up strategy \cite{shen2023triplet} for the first 500 iterations. We train the model on the DOTAV1.0 \cite{DOTA_dataset} dataset for 12 epochs and on the HRSC2016 \cite{HRSC2016} dataset for 72 epochs. Weight decay and momentum are 0.0001 and 0.9, respectively. 
\begin{table}[t!]
\centering
\caption{Ablation study on the interaction methods within the contextual interaction module (CIM).
All experiments were conducted based on adding the SPIEM.}
\begin{tabular}{cc|cc|cc|ccc}
\hline
  \hline
 \multicolumn{2}{c|}{\multirow{3}*{Methods}} & \multicolumn{4}{c|}{CIM} & \multicolumn{3}{c}{Metric}  \\
  \cline{3-9}
    & &\multicolumn{2}{c|}{\multirow{2}*{$\bf{CI}$}}  &\multicolumn{2}{c|}{$\bf{SI}$}
   &\multirow{2}*{AP$_{50}$}  & \multirow{2}*{AP$_{75}$} & \multicolumn{1}{c}{\multirow{2}*{mAP}}  \\
  \cline{5-6}
    & &  & &$\bf{NI}$  & $\bf{LI}$  &\multicolumn{3}{c}{}\\
  \hline
   \multicolumn{2}{c|}{Baseline + SPIEM}  & \multicolumn{2}{>{\color{gray}}c|}{\ding{55}}    &\color{gray}\ding{55}   &\color{gray}\ding{55} & 89.1 & 66.3 & 56.1 \\
  
   \multicolumn{2}{c|}{+ $\bf{CI}$ + $\bf{SI}$ (only $\bf{NI}$)} &\multicolumn{2}{c|}{\ding{51}}    &\ding{51}       &\color{gray}\ding{55}       & 89.9 & 64.3 & 57.2 \\
  
   \multicolumn{2}{c|}{+ $\bf{CI}$ + $\bf{SI}$ (only $\bf{LI}$)}  &\multicolumn{2}{c|}{\ding{51}}    &\color{gray}\ding{55} &\ding{51} & 89.9 & 66.5 & 57.0  \\
    \hline 
   \multicolumn{2}{c|}{LR-FPN (Ours)}      &\multicolumn{2}{c|}{\ding{51}}&\ding{51}     &\ding{51}    &{\bfseries 90.4} &{\bfseries 68.9} &{\bfseries 59.5}\\
  \hline
    \hline
\end{tabular}
\label{tab:CIM ablation}
\end{table}

\subsection{Comparison with State-of-the-art Methods } 
\subsubsection{Comparisons on DOTAV1.0}
{\color{black} From Table \ref{tab:sota}, It can be observed that LR-FPN excels across all objective indices. Compared to Faster-RCNN, which also employs ResNet50 as its backbone network, the proposed LR-FPN maintains strong competitiveness. This suggests that enhancing the interaction in the information propagation process, facilitating the fusion of features from multiple scales, can lead to a better understanding of features. Our mAP metric significantly outperforms that of PVANet-FFN~\cite{PVANet-FFN}, which uses PVANet as the backbone network, indicating that extracting positional and saliency information from low-level feature maps to maximize the enhancement and retention of location information can significantly improve the performance. }

\subsubsection{Comparisons on HRSC2016 \cite{HRSC2016}}
{\color{black}As shown in Table \ref{table:HRSC2016}, we present the performance comparison of LR-FPN with other state-of-the-art methods on the HRSC2016 dataset \cite{HRSC2016}. Experimental results indicate that LR-FPN excels in terms of mAP, surpassing the baseline (R$^3$Det \cite{R3det}) by 2.2\%. The observed improvement in the performance of LR-FPN on the HRSC2016 \cite{HRSC2016} dataset aligns with its performance on the DOTAV1.0 \cite{DOTA_dataset} dataset, suggesting that the results generated by LR-FPN are more stable across various scenarios in remote sensing datasets. }

\subsection{Ablation Studies and Analysis}  
In the previous subsection, we have shown the superiority of LR-FPN by comparing it to state-of-the-art methods.
In what follows, we comprehensively analyze intrinsic factors that lead to LR-FPN’s superiority in DOTAV1.0 and HRSC2016.

\textbf{ The role of LR-FPN.} 
 As shown in the first group of Table \ref{tab:hrsc ablation}, after using the SPIEM independently, the model achieves 0.9\% improvement on both AP$_{50}$ and AP$_{75}$, respectively. The second group leveraging CIM independently reflects that AP$_{50}$ and AP$_{75}$ achieve growth of 1.5\% and 0.7\%. 
The effectiveness of SPIEM and CIM are verified. This indicates that SPIEM excels in extracting precise positional and saliency information, maximizing the enhancement and preservation of location details. On the other hand, CIM facilitates enriched contextual interaction by considering spatial and channel dimensions. 

Moreover, the joint effect between modules is illustrated in the 7$^{th}$ group of Table \ref{tab:hrsc ablation}. Simultaneously introducing SPIEM and CIM, AP$_{50}$, AP$_{75}$ and mAP increase by 2.2\%, 3.5\% and 3.6\%, respectively. The results demonstrate that SPIEM optimizes the preservation of precise location information of the target, while CIM seamlessly and effectively integrates and processes the information, which enhances the comprehensive ability of our model.

\begin{figure}[t]
    \centering
    \includegraphics[width=0.9\linewidth]{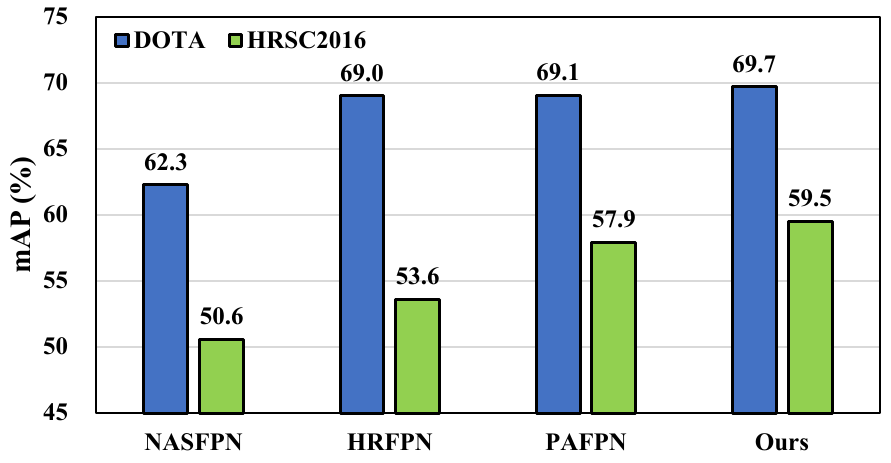}
    \vspace{-0.2cm}
     \caption{The comparison with various FPNs.}
\label{fig: fpn sota}
\end{figure}

\textbf{Comparisons with different FPNs.} In order to further validate the effectiveness of LR-FPN, our comparison includes advanced FPN variants, namely NASFPN \cite{nasfpn}, HRFPN \cite{hrnet}, and PAFPN \cite{pafpn}.
 Fig. \ref{fig: fpn sota} depicts the detection outcomes achieved through the utilization of advanced feature pyramid networks. 
In the DOTAV1.0 \cite{DOTA_dataset}, our LR-FPN demonstrates superior performance over the suboptimal PAFPN \cite{pafpn}, exhibiting a 0.6\% improvement in mAP. Similarly, in the HRSC2016 \cite{HRSC2016}, our LR-FPN achieves a substantial 1.6\% increase in mAP when compared to the PAFPN \cite{pafpn}, showcasing its superior performance across multiple benchmarks. Benefited by the successful extraction of shallow positional information and the effective interaction of fine-grained context information, our network demonstrates exceptional applicability in remote sensing scenes. This further consolidates its effectiveness within this specific domain.

\par
\textbf{The effectiveness of variant SPIEM.}
The 3$^{rd}$ and 4$^{th}$ groups in Table \ref{tab:hrsc ablation} show the result of variant SPIEM. It is observed that incorporating either the SP or PP method in CIM only results in marginal improvements in the AP${_{50}}$ metric. However, when considering the AP$_{75}$ metric, the addition of the SP method increased the metric by 1.5\% and the addition of the PP method increased it by 0.9\%. This observation suggests that both the SP and PP methods are capable of extracting valuable localization information and salient features, thereby maximizing the enhancement and retention of location information when there is a requirement for more precise localization. Significantly, when the SP and PP methods are combined, their effects become more prominent and can lead to a more substantial improvement in performance.

\begin{figure}[t]
    \centering
    \includegraphics[width=0.9\linewidth]{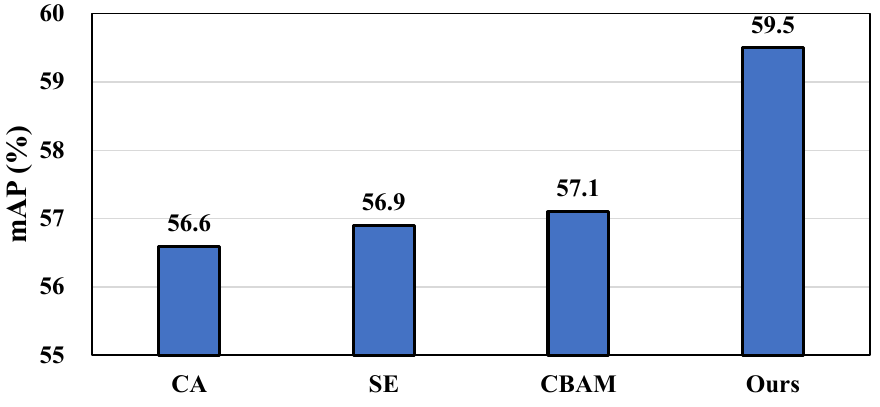}
     \caption{The comparison with variants of CIM.}
\label{fig: variant of CIM}
\end{figure}

 \begin{figure*}[t]
    \centering
    \includegraphics[width=0.95\linewidth]{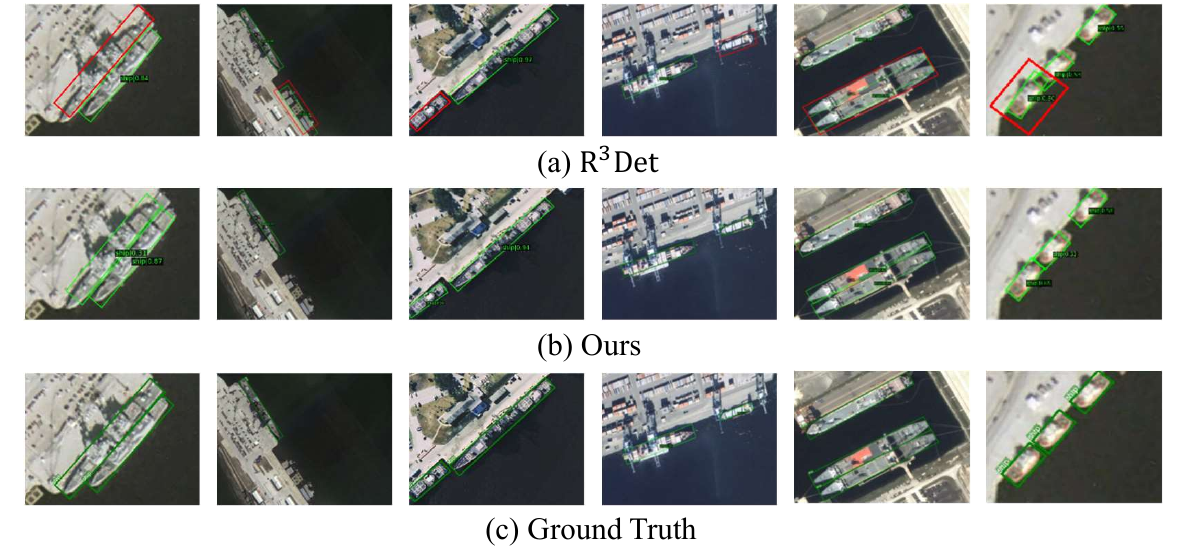}
    \caption{Visualizations of detection results on HRSC2016 \cite{HRSC2016}. Failures are marked by red boxes.}
    \label{fig:vison_hrsc}
\end{figure*}

\par
\textbf{The impact of variant CIM.} To delve deeper into CIM's functionality and assess its performance, we design a series of comprehensive experiments. The 5$^{th}$ and 6$^{th}$ group in Table \ref{tab:hrsc ablation} unequivocally demonstrates that our spatial and channel interaction blocks effectively promote information interaction. What's more, Table \ref{tab:CIM ablation} presents the outcomes obtained from employing various interaction methods within CIM. Both local and non-local interaction blocks demonstrate improvement in the metrics, and their combined effect is more pronounced. This also indicates that our CIM not only facilitates interaction among local information but also establishes long-range dependencies. Additionally, we conduct a comparison between our interaction method and advanced attention methods, including CA \cite{CA}, SE \cite{SE}, and CBAM \cite{CBAM}. The results and findings of this comparison are illustrated in Fig. \ref{fig: variant of CIM}. Through this analysis, we aim to evaluate the performance and effectiveness of our interaction method with these state-of-the-art attention methods.
 Regarding mAP, CIM shows superior performance to the state-of-the-art CBAM \cite{CBAM}, achieving a higher accuracy of 2.4\%. This result proves that our contextual interaction method effectively facilitates fine-grained context information while accounting for long-range dependencies, improving object coverage and area.

\subsection{Visualization}  
From Fig. \ref{fig:vison_hrsc}, we present three sets of results on the HRSC2016 \cite{HRSC2016} dataset. They include the outputs of R$^3$Det (Baseline) \cite{R3det}, the outputs of LR-FPN  (Ours), and the ground truth (Ground Truth). 
Compared to the first row (R$^3$Det), we observe that our model exhibits a higher level of precision in orienting the ships. In addition, in terms of both false positives and false negatives detection, our model outperforms the baseline. It can be seen from this that our SPIEM undoubtedly extracts effect location information, while CIM facilitates contextual interaction, thus enhancing the overall capabilities of our model. Thanks to our modules, the fusion and learning of multi-scale feature information show great performance, which contributes in particular to detecting challenging multi-scale objects, especially in the remote sensing field.

\section{Conclusion}\label{sec:con} 
This study addressed the shortcomings of existing feature pyramid networks in remote sensing target detection, specifically their disregard for low-level positional information and fine-grained context interaction. We introduced a novel location-refined feature pyramid network (LR-FPN) that enhances shallow positional information extraction and promotes fine-grained context interaction. The LR-FPN, equipped with a shallow position information extraction module (SPIEM) and a contextual interaction module (CIM), effectively harnessed robust location information. We also implemented a local and non-local interaction strategy for superior saliency information retention. The LR-FPN can be incorporated into standard object detection frameworks, significantly boosting performance.
{\textbf{In the future.}} Despite LR-FPN's top-tier results on two prevalent remote sensing datasets using a CNN-based architecture, its performance within a transformer-based framework remains unverified. We will further investigate the potential of implementing our approach within a transformer-based architecture.

\bibliographystyle{IEEEtran}
{\small \bibliography{ref} }

\end{document}